\documentclass[10pt,twocolumn,letterpaper]{article}

\usepackage{cvpr}              

\usepackage{graphicx}
\usepackage{amsmath}
\usepackage{amssymb}
\usepackage{booktabs}
\usepackage{enumitem}

\usepackage[pagebackref,breaklinks,colorlinks]{hyperref}

\usepackage[capitalize]{cleveref}
\crefname{section}{Sec.}{Secs.}
\Crefname{section}{Section}{Sections}
\Crefname{table}{Table}{Tables}
\crefname{table}{Tab.}{Tabs.}

\def\cvprPaperID{5157} 
\def\confName{CVPR}
\def\confYear{2023}

\usepackage[table]{xcolor}

\newcommand{\boldparagraph}[1]{\vspace{0.0cm}\noindent{\bf #1}}

\begin{document}

\title{PlaneDepth: Self-supervised Depth Estimation via Orthogonal Planes}

\author{Ruoyu Wang$^1$ \hspace{1cm} Zehao Yu$^2$ \hspace{1cm} Shenghua Gao$^{1,3,4}$\thanks{Corresponding Author.}\\
$^1$ShanghaiTech University \hspace{1cm} $^2$University of Tübingen\\
$^3$Shanghai Engineering Research Center of Intelligent Vision and Imaging\\
$^4$Shanghai Engineering Research Center of Energy Efficient and Custom AI IC\\
{\tt\small \{wangry3, gaoshh\}@shanghaitech.edu.cn, zehao.yu@uni-tuebingen.de}
}

\maketitle

\begin{abstract}
   Multiple near frontal-parallel planes based depth representation demonstrated impressive results in self-supervised monocular depth estimation (MDE). Whereas, such a representation would cause the discontinuity of the ground as it is perpendicular to the frontal-parallel planes, which is detrimental to the identification of drivable space in autonomous driving. In this paper, we propose the PlaneDepth, a novel orthogonal planes based presentation, including vertical planes and ground planes. PlaneDepth estimates the depth distribution using a Laplacian Mixture Model based on orthogonal planes for an input image. These planes are used to synthesize a reference view to provide the self-supervision signal. Further, we find that the widely used resizing and cropping data augmentation breaks the orthogonality assumptions, leading to inferior plane predictions. We address this problem by explicitly constructing the resizing cropping transformation to rectify the predefined planes and predicted camera pose. Moreover, we propose an augmented self-distillation loss supervised with a bilateral occlusion mask to boost the robustness of orthogonal planes representation for occlusions. Thanks to our orthogonal planes representation, we can extract the ground plane in an unsupervised manner, which is important for autonomous driving. Extensive experiments on the KITTI dataset demonstrate the effectiveness and efficiency of our method. The code is available at \url{https://github.com/svip-lab/PlaneDepth}.
\end{abstract}

\section{Introduction}
Monocular depth estimation (MDE) is an important task in computer vision and it has tremendous potential applications, such as autonomous driving. However, the expensive data and labels acquisition process restricts the data scale in supervised MDE \cite{gur2019single, luo2018single, fu2018deep, li2022binsformer, bhat2021adabins, bhat2022localbins}. Thus, researchers turn to solve the data constraints in supervised MDE with the self-supervised MDE framework by leveraging videos or stereo image pairs. 

\begin{figure*}[!t]
\centering
\includegraphics[width=\linewidth]{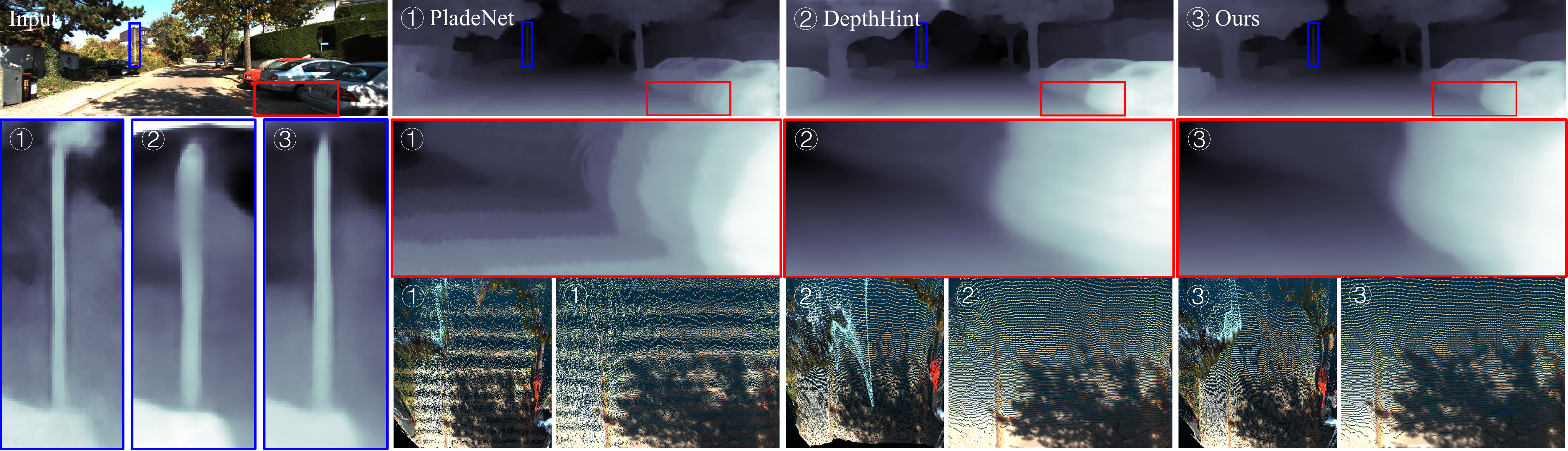}
\caption{\textbf{Monocular Depth Estimation Results}. The zoomed-in visualizations and bird-eye-view colored point cloud show that our method can predict continuous depth in the ground region while preserving the sharp edges of objects. Compared with PladeNet~\cite{gonzalez2021plade}, our prediction has smoother depth in the ground region. Compared with DepthHint~\cite{watson2019self}, our method addresses the occlusion problem, which can be seen in figure that our prediction have eliminated depth artifacts in the edges of street lights and cars.
}
\label{fig:top}
\end{figure*}

Most of the early works in self-supervised MDE leverage a regression module to estimate pixel-wise depth map~\cite{godard2017unsupervised, gordon2019depth, godard2019digging, watson2019self, shu2020feature, peng2021excavating} and warp the reference view image to the target view based on the estimated depth. Then, a photometric consistency loss is used to guide the learning of the depth regression module. However, these methods usually encounter the local minimum issue because of the locality of bilinear interpolation on the reference view. To avoid this issue, rather than using simple depth regression, multiple frontal-parallel planes based depth representation is introduced where depth space is divided into a fixed number of frontal-parallel planes, and the depth network learns to classify which predefined plane each pixel belongs to~\cite{gonzalezbello2020forget, gonzalez2021plade}. It has been shown that such representation could produce much sharper depth on the edges of the object. However, they are insufficient to represent the ground because the ground plane is perpendicular to these predefined frontal-parallel planes. As shown in~\cref{fig:top}, such vertical depth planes only solution would lead to discontinuity on the ground, which is obviously detrimental to the identification of drivable space in autonomous driving. Further, photometric consistency loss is applied to the weighted composition of each plane-warped image, which is sub-optimal as the combination of different weights may lead to exactly the same color image~\cite{peng2022rethinking}, resulting in ambiguous solutions for depth plane classification. 

Considering the ground is perpendicular to the frontal-parallel planes, in this paper, we propose to leverage orthogonal planes to represent the scene where the ground planes favor the depth estimation in the ground region. Further, we propose to model the depth as a mixture of Laplace distributions of orthogonal planes~\cite{tosi2021smd}, where each Laplacian is centered at one plane. We compute the photometric consistency loss independently on the color image warped by each plane, resulting in a more deterministic and less ambiguous optimization objective compared with the weighted composition strategy mentioned above~\cite{gonzalezbello2020forget, gonzalez2021plade}, consequently leading to better depth estimation results, as shown in ~\cref{fig:top}. Moreover, we can extract the ground plane in an unsupervised manner thanks to our orthogonal planes representation.

Resizing and cropping are widely used as a data augmentation strategy in the stereo training setting of MDE \cite{gonzalezbello2020forget, gonzalez2021plade}. However, it would destroy the orthogonality of our predefined planes, leading to inferior plane distributions estimation. To remedy this issue, in this paper, we deeply analyze the effects of resizing and cropping on the world coordinates system. We explicitly compute the resizing and cropping transformation and use it to rectify the predefined planes and the predicted camera rotation, which eases the learning of plane distributions. We further propose to use neural positional encoding (NPE) for the resizing and cropping parameter and incorporate it into our PlaneDepth network, which improves the robustness of network training. 

The self-distillation strategy is commonly used to solve occlusion problems \cite{gonzalezbello2020forget, gonzalez2021plade} and improve the depth prediction results \cite{peng2021excavating}. Post-processing is widely used to improve the final prediction~\cite{godard2017unsupervised}, which can be used naturally to generate more accurate self-distillation labels. In this paper, we propose to combine post-processing with self-distillation by using a bilateral occlusion mask to generate more accurate supervision of network training, which improves both accuracy and efficiency of our method. 

We summarize our contributions as follows:
\begin{enumerate}[nosep]
   \item We propose the PlaneDepth, a novel orthogonal plane-based monocular depth estimation network, which favours the representation of both vertical objects and ground. Such representation leads to a much smoother depth for ground regions and would facilitate the identification of drivable regions. 
 \item The depth within the scene is modeled by a mixture of Laplacian distributions, and the depth classification problem is cast as the optimizing the mixture of Laplace distribution, which avoids the ambiguity in color expectation based depth estimation and leads to more stable depth estimation.
  \item An orthogonality-preserved data augmentation strategy is proposed, which improves the robustness of network training. 
  \item We combine post-processing with self-distillation by our augmented self-distillation, which improves both efficiency and accuracy.
\end{enumerate}

\begin{figure*}[!t]
\centering
\includegraphics[width=\linewidth]{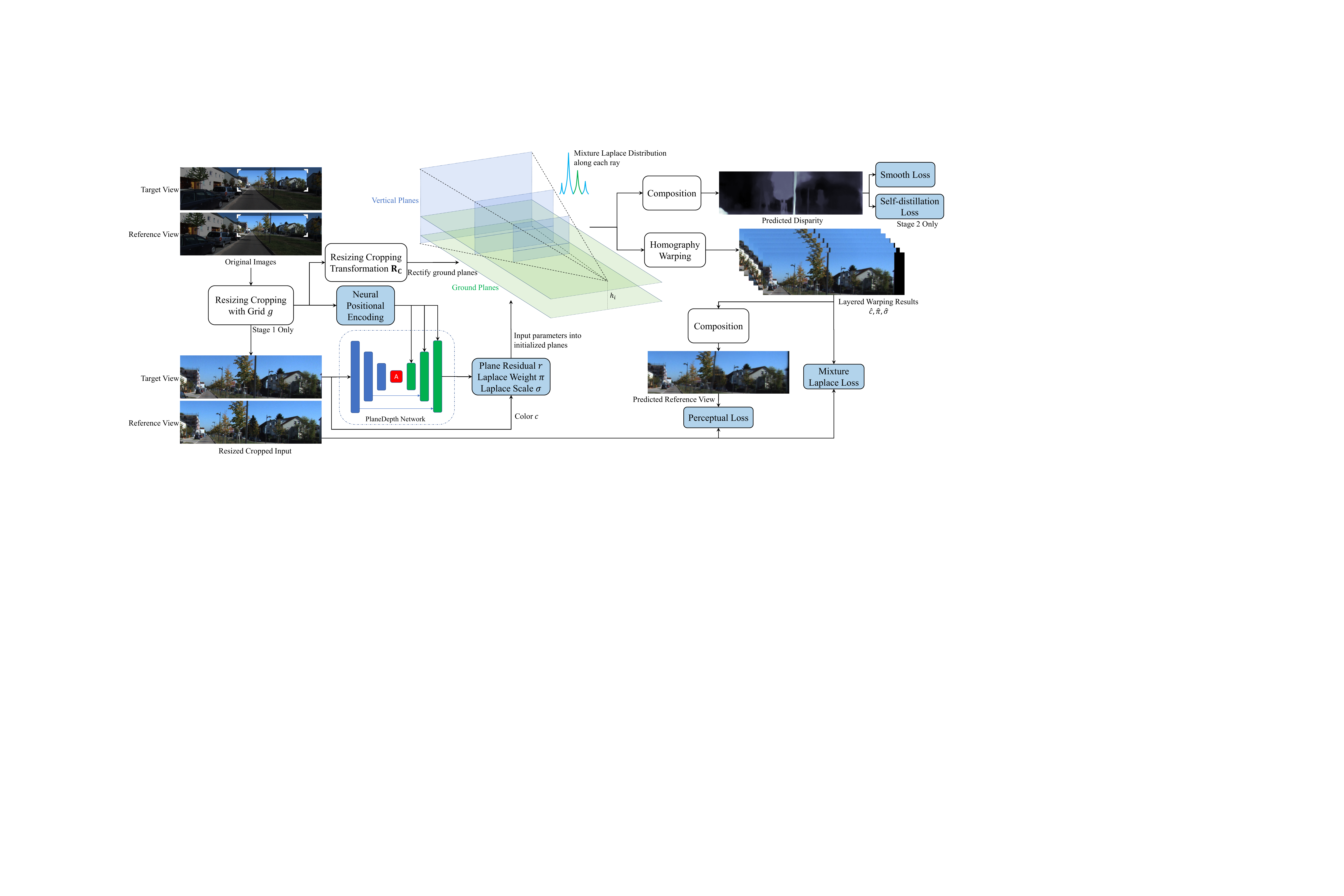}
\caption{\textbf{PlaneDepth Pipeline}.
We model the depth as a mixture of Laplace distribution of orthogonal planes for self-supervised depth estimation. Our PlaneDepth network predicts distribution parameters for these planes, which together with the target view image are warped to the reference view. The plane-warped images are composed to synthesize a predicted reference view which is supervised with perceptual loss. We further supervise each individual plane-warped image with the photometric consistency loss between the reference view weighted by the Laplace distributions. We address the non-orthogonality issue due to resizing and cropping by explicitly rectifying predefined planes and neural positional encoding from the resizing cropping grid. A self-distillation loss is used to further improve the performance. We use double-channel ResNet18\cite{he2016deep} following Godard \etal~\cite{godard2019digging} and add NPE to predict the relative pose of two views in our monocular training experiments. The blue and white boxes indicate the differentiable variables and the hard operations, respectively.
}
\label{fig:pipeline}
\end{figure*}

\section{Related Works}

\subsection{Self-supervised Depth Estimation}
Self-supervised learning of depth estimation from video sequences or stereo image pairs has been proposed to ease the demand for large-scale labeled data. Since the seminal work by  Zhou~\etal~\cite{zhou2017unsupervised} that demonstrated inspiring results by jointly optimizing depth and pose networks with an image reconstruction loss. Many following work further improve the performance through occlusion modeling~\cite{poggi2018learning, wong2019bilateral, godard2019digging, ramamonjisoa2020predicting}, addressing scale ambiguity~\cite{zhang2022towards}, modeling moving objects~\cite{gordon2019depth, klingner2020self, bello2022positional}, enforcing depth consistency~\cite{godard2017unsupervised, he2022ra}, designing better network architectures~\cite{pillai2019superdepth, han2022brnet, zhou2022self}, incorporating additional semantic information~\cite{chen2019towards, jung2021fine, chen2022self, ma2022towards, zhu2020edge}, exploiting relative object size cues~\cite{gonzalezbello2020forget, peng2021excavating}, or extending to indoor scenes~\cite{yu2020p, li2021mine, ji2021monoindoor}. In this paper, we propose to utilize orthogonal planes for better depth representation in driving scenarios and demonstrate improved performance with orthogonal planes.

\subsection{Plane-based Scene Representation}
Benefiting from the simplicity of planes and the efficiency of differentiable homography warping, multi-plane image (MPI) is widely used in novel view synthesis~\cite{zhou2018stereo,gur2019single, li2021mine}. The idea is also introduced in the depth estimation~\cite{gonzalezbello2020forget, gonzalez2021plade} recently and has been shown effective in self-supervised depth estimation since it addresses the local minima problem caused by local searching in the reference view. Zhou and Dong~\cite{zhou2022learning} further address the depth discontinuity problem by regressing a depth residual for each plane. However, these MPI-based networks will predict vertical-prone depth in the ground region which is the dominant region in the outdoor driving scenario. In contrast, we propose an orthogonal planes based representation that can model continuous depth and free space in the ground region while preserving the sharp edges of objects.

\subsection{Self-distillation for Depth Estimation}

Self-distillation is a technique commonly used in self-supervised depth estimation that involves using a pretrained teacher network to generate labels for training a better student network. For instance, Peng~\etal~\cite{pilzer2019refine,peng2021excavating} synthesized multiple predictions from teacher network to generate pseudo labels, which are then used to train the student network. Self-distillation can also be used to reduce the uncertainty of the prediction~\cite{bello2021self, xu2021digging}, or solve the artifact problem caused by occlusion~\cite{gonzalezbello2020forget,zhou2022learning}. Inspired by these works and the widely-used post-processing (PP) technique, which inputs both an original image and its flipped version into a depth network and averages the predictions for more accurate depth estimation, we propose to utilize a bilateral occlusion mask into PP technique to synthesis pseudo labels which is used as an additional distillation loss to further improve our depth network.

\section{Methods}

Previous works \cite{gonzalezbello2020forget, gonzalez2021plade} use planes parallel to the camera to represent the scene which would lead to a discontinuity of the ground. In this paper, we propose to model the depth distribution with a mixture of Laplacian based on a set of orthogonal planes consisting of the vertical and ground planes. Then orthogonality-preserved data augmentation is introduced by taking the resized cropping transformation and Neural Positional Encoding. We also propose an augmented self-distillation with the supervision of the bilateral occlusion mask to improve the robustness of our PlaneDepth to occlusions. 

\subsection{Orthogonal Planes based Representation}
\label{sec:plane_prediction}
To improve the capability of the network in predicting the depth for both ground and objects at a distance, we propose to represent the scene with a set of vertical planes as well as ground planes, which are orthogonal. 

\boldparagraph{Plane Definition}. Take the target view as the origin, the planes are defined as:
\begin{equation}
\label{planes}
\mathbf{n}_i^T\mathbf{w}-\delta_{i}=0,\quad i=0,1,\dots,N-1
\end{equation}
where $\mathbf{n}$ is the normal of the plane, $\mathbf{w}$ is a point in the world coordinate system, $\delta_i$ is the distance from the $i$-th plane to the origin, and $N$ is the number of planes. 

\boldparagraph{Vertical Planes}. 
Following~\cite{gonzalezbello2020forget}, we initialize $N^v$ vertical planes with $\mathbf{n}^v=[0\ 0\ 1]^T$, and sample $\delta_i^v$ in exponential disparity space:
\begin{equation}
\label{d_space}
\delta_i^v = \frac{Bf_x}{d_i^v}\hspace{1cm} 
d_{i}^v = d_\text{max} (\frac{d_\text{min}}{d_\text{max}})^\frac{i+r_{i}^v}{N_v-1}
\end{equation}
where $B$ is the baseline of the stereo pair and $f_x$ is the horizontal focal length, $d_\text{min}$ and $d_\text{max}$ are the minimum and maximum disparity, respectively. $r_i^v$ are predicted by the network to account for discretization errors.

\boldparagraph{Ground Planes}. Similarly, we initialize $N^g$ ground planes with $\mathbf{n}^g=[0\ 1\ 0]^T$ and different $\delta_i^g$ linearly in camera height space:
\begin{equation}
\label{h_space}
\delta_{i}^g=h_{\text{min}}+\frac{i+r_{i}^g}{N_{g}-1}(h_{\text{max}}-h_{\text{min}}),
\end{equation}
where $h_{min}$ and $h_{max}$ are the minimum and maximum height of the ground planes, respectively. Again, $r_i^g$ are predicted by the network to account for discretization errors.

\boldparagraph{Depth from Orthogonal Planes}. We model the depth as a mixture of Laplace distributions of orthogonal planes, with each Laplacian centered at one plane. Therefore, PlaneDepth also predicts weights $\pi_i$ and scales $\sigma_i$ for each Laplace distribution at each pixel, where $\pi_i$ is normalized by a softmax function $\mathcal{S}$ of logit $l_i$. Plane residuals $r_i^v$ and $r_i^g$ are shared by all pixels, and each pixel has its own parameters for mixture Laplace distributions. We omit the subscript of pixel for simplicity. Now we can compute the depth $\hat{D}$ for each pixel via composition:
\begin{equation}
\label{pred_depth}
\hat{D}=\frac{1}{Z}\sum^{N-1}_{i=0}p_iD_i \hspace{1cm} p_i = \sum_{j=0}^{N-1} \frac{\pi_{j}e^{\frac{-|D_i - D_j|}{\sigma_j}}}{2\sigma_j}
\end{equation}
where $p_i$ is the probability that the pixel belongs to $i$-th plane, $Z=\sum_{i=0}^{N-1} p_i$ is a normalization constant and $D_i$ is rendered depth of $i$-th plane.

\boldparagraph{Homography Warping}. Given relative camera pose $\mathbf{R}$ and $\mathbf{t}$, either from calibrated stereo pairs or predicted by a pose network, between the target image and the reference image, we construct homography mappings $\mathbf{H}_i$  for the $i$-th plane:
\begin{equation}
\label{homography}
\mathbf{H}_i = \mathbf{K}(\mathbf{R}+\frac{1}{\delta_i}(\mathbf{t}\mathbf{n}_i^T))\mathbf{K}^{-1},
\end{equation}
where $\mathbf{K}$ is the camera intrinsic matrix. Following~\cite{gonzalezbello2020forget, gonzalez2021plade, zhou2022self}, we warp the target image $\mathbf{I}_t$ together with pixel-wise plane distributions $\pi_i$, $\sigma_i$ to the reference view via bilinear sampling $\mathcal{G}$: 
\begin{equation}
\hat{\mathbf{I}}_i, \hat{l}_i, \hat{\sigma}_i = \mathcal{G}(\mathbf{H}_i, \mathbf{I}_t, l_i, \sigma_i) \hspace{0.5cm}
\hat{\pi}=\mathcal{S}([\hat{l}_i]_{i=0}^{N-1}).
\end{equation}
Note that our warped weights $\hat{\pi}$ is obtained by warping logit $l$ followed by softmax $\mathcal{S}$~\cite{gonzalezbello2020forget}.
Then we can synthesis a reference image $\hat{\mathbf{I}}_r$ via composition:
\begin{equation}
\hat{\mathbf{I}}_r = \frac{1}{\hat{Z}}\sum^{N-1}_{i=0}\hat{p}_{i}\hat{\mathbf{I}}_{i} \hspace{1cm} \hat{p}_i = \sum_{j=0}^{N-1} \frac{\hat{\pi}_{j}e^{\frac{-|D^r_i - D^r_j|}{\hat{\sigma}_j}}}{2\hat{\sigma}_j},
\end{equation}
where $\hat{Z}=\sum_{i=0}^{N-1} \hat{p}_i$ and $D_i^r$ is rendered depth of the $i$-th plane in the reference view.

\boldparagraph{Optimization}. We use perceptual loss~\cite{johnson2016perceptual} to constrain the feature-level similarity of the synthesis reference view:
\begin{equation}
\label{pc_loss}
\mathcal{L}_{\text{pc}} = ||\phi_l(\mathbf{I}_r)-\phi_l(\hat{\mathbf{I}}_r)||^2_2,
\end{equation}
where $\phi_l$ is the first $l$ maxpool layers of a VGG19\cite{simonyan2014very} pretrained on the ImageNet\cite{deng2009imagenet}. Further, we propose a novel photometric consistency based on Laplace distributions:
\begin{equation}
\label{ph_loss}
\mathcal{L}_{\text{MLL}} = -\log\sum_{i=0}^{N-1} \frac{\hat{\pi}_{i}e^{\frac{-\frac{1}{3}||\mathbf{I}_r - \hat{\mathbf{I}}_{i}||_1}{\hat{\sigma}_{i}}}}{2\hat{\sigma}_{i}}.
\end{equation}
Compared to $L_1$ loss which supervise the final composite image $\hat{\mathbf{I}}_t$, our $\mathcal{L}_{\text{MLL}}$ supervise each plane warped image $\hat{\mathbf{I}}_i$, resulting in a more deterministic optimization objective.

We also use smooth loss $\mathcal{L}_{\text{ds}}$ for depth map $\hat{D}$ as it has been shown effective in many recent works\cite{godard2019digging, watson2019self, shu2020feature, gonzalezbello2020forget, gonzalez2021plade, li2021mine}. 
Therefore, our final loss is:
\begin{equation}
\label{loss}
\mathcal{L}=\mathcal{L}_{\text{MLL}} + \lambda_{1}\mathcal{L}_{\text{pc}} + \lambda_{2}\mathcal{L}_{\text{ds}}
\end{equation}
which is averaged over pixels, views and batches. $\lambda_{1}$ and $\lambda_{2}$ are hyper-parameters to balance different loss terms.

\subsection{Orthogonality-preserved Data Augmentation}
\label{sec:resize_crop}
\begin{figure}[t]
\centering
\includegraphics[width=\linewidth]{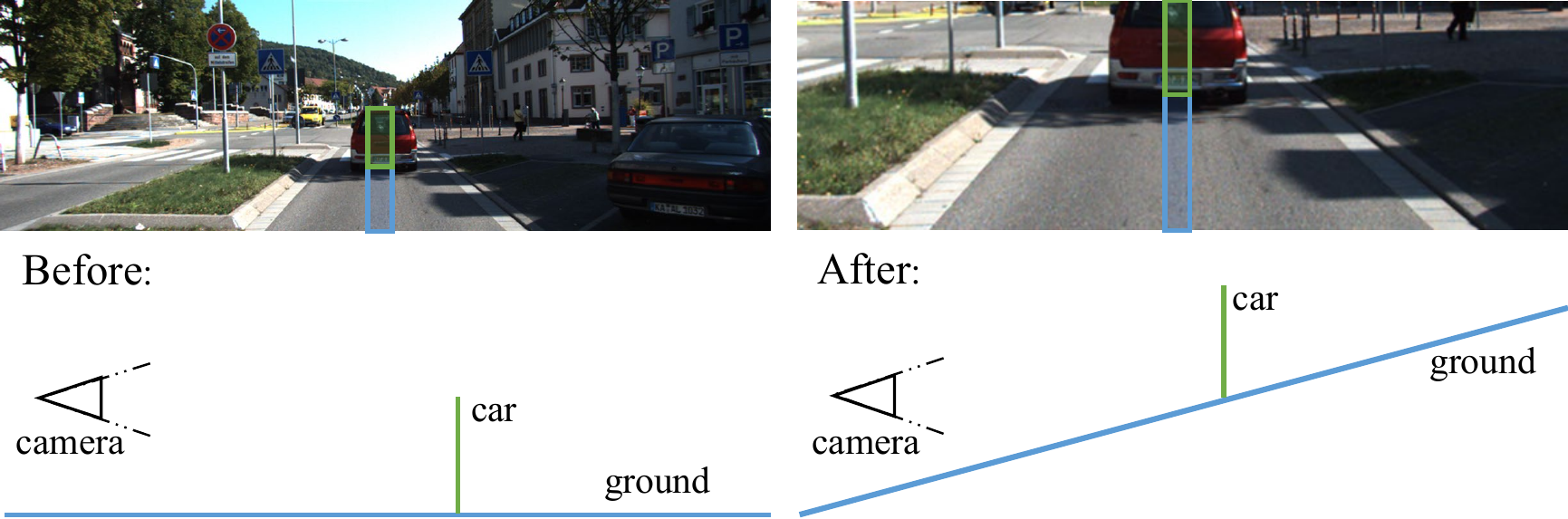}
\caption{Resizing cropping causes planes to be non-orthogonal.}
\label{fig:relative_size}
\end{figure}

Resizing and cropping is a widely used data augmentation strategy, which encourages the MDE to exploit relative size cue~\cite{bello2022positional}. However, when the crop does not occur at the center height of the image, the ground will tilt, as shown in \cref{fig:relative_size}, and contradict our orthogonal planes assumptions. 
Consequently, our PlaneDepth network struggles to predict accurate parameters for such augmented data. We address this issue by calculating the transformation matrix of the augmented data explicitly. 

We assume that when the size of the object becomes $f_s$ times larger, its depth is reduced by a factor $f_s$. Given an image corresponding to a camera intrinsic with focal length ($f_x,\ f_y$) and principal point ($c_x,\ c_y$), when it is scaled with a factor $f_s$ and cropped with a grid centered at ($p_x,\ p_y$), the resizing and cropping lead to a transformation $\mathbf{R_C}$ which transforms the original world coordinates $w$ to the new coordinates $\tilde{w}$ originated at the augmented image:
\begin{equation}
\label{Rc}
\tilde{\mathbf{w}} = \mathbf{R_C}\mathbf{w} \hspace{1cm} \mathbf{R_C}=
\begin{bmatrix}
1 &   & \frac{c_x-p_x}{f_x} \\ 
  & 1 & \frac{c_y-p_y}{f_y} \\
  &   & f_s
\end{bmatrix}.
\end{equation}

Similarly, the normal $\mathbf{n}$ and distance $\delta$ of planes in the augmented image are also transformed as:
\begin{equation}
\label{ground_planes_rc}
\tilde{\mathbf{n}} = \frac{\mathbf{R_C}^{-T}}{||\mathbf{R_C}^{-T}\mathbf{n}||}\mathbf{n}\hspace{1cm} \tilde{\delta} = \frac{\delta}{||\mathbf{R_C}^{-T}\mathbf{n}||}.
\end{equation}
As a result, the vertical planes $\tilde{\mathbf{n}}^v$ and $\tilde{\mathbf{n}}^g$ after transformation are not orthogonal\footnote{More details are provided in the supplementary material.}. To maintain the orthogonality of our predefined planes, we use \cref{ground_planes_rc} to rectify ground planes during training such that they are always perpendicular to the ground after resizing and cropping.

Further, after resizing and cropping augmentation, it is desirable that the networks would predict the depth, disparity and pose as~\footnote{The derivation of \cref{Rc}, \cref{ground_planes_rc} and \cref{dn_Dn_Rn_tn} are provided in the supplementary material.}:
\begin{equation}
\label{dn_Dn_Rn_tn}
\tilde{d}=\frac{d}{f_s},\ \widetilde{D}=f_s D,\ \widetilde{\mathbf{R}}=\mathbf{R_C}\mathbf{R}\mathbf{R_C}^{-1},\ \tilde{\mathbf{t}}=\mathbf{R_C}\mathbf{t}
\end{equation}
However, $\widetilde{\mathbf{R}}$ is not orthogonal and hence not a canonical rotation matrix. Therefore, $\widetilde{\mathbf{R}}$ cannot be predicted with 3 degrees of freedom as existing pose estimation methods~\cite{xu2021digging}. In order to be compatible with these methods, we explicitly rectify the rotation matrix $\mathbf{R}$ predicted by the pose network to $\widetilde{\mathbf{R}}$ according to \cref{dn_Dn_Rn_tn}, which ease the learning of the pose network.

\boldparagraph{Neural Positional Encoding}. After adjusting the planes and camera rotation, the network still has difficulty in segmenting the ground because it needs to predict the degree of inclination caused by resize cropping before predicting the ground height $\delta^g$. In order to help the network to better segment the ground, we encode the $\mathbf{g}$ with the neural positional encoding (NPE)~\cite{gonzalez2021plade}, where the grid $\mathbf{g} \in [-1,1]^2$ is represented as the relative position of each pixel in the original image, after the resizing and cropping transformation. Since our encoder is designed to extract image features based on the relative sizes of objects, we only feed the NPE of the grid into each layer of the decoder, as shown in \cref{fig:pipeline}. Similarly for the pose network, in order to avoid the network predicting changes from $\widetilde{\mathbf{R}}$ to $\mathbf{R}$, we also feed NPE of $\mathbf{g}$ into the pose network.

\subsection{Augmented Self-distillation}
\label{sec:self_distillation}
\begin{figure}[!t]
\centering
\includegraphics[width=\linewidth]{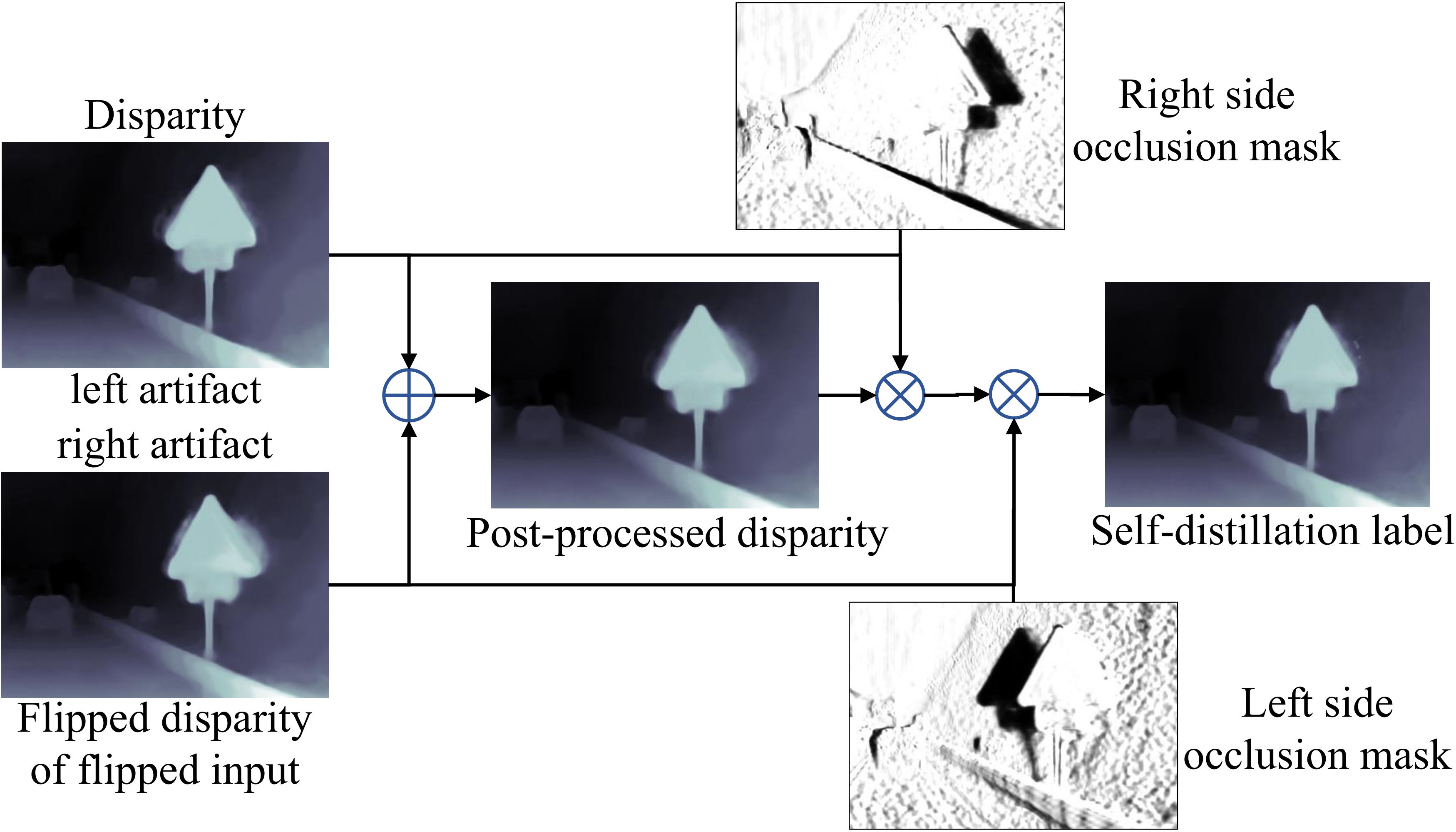}
\caption{\textbf{Obtain the self-distillation label}. Our final self-distillation labels are obtained by filtering the artifacts on both sides of the post-processed disparity map using the corresponding side mask and accurate disparity respectively.}
\label{fig:self_distillation}
\end{figure}
Self-distillation is proven to be effective for getting rid of occlusion effects~\cite{gonzalezbello2020forget, gonzalez2021plade, zhou2022self}. Since pixels that are not visible in the right view exist only on the left of objects in the left view, which means that the photometric matching failure caused by occlusion only occurs on the left side of the object in the left view. Hence, a network that takes only the left view as input and finds matchings in the right view will produce artifacts to the left of objects. 
Gonzalez \etal~\cite{gonzalezbello2020forget} exploits this property to generate pseudo labels for self-distillation with a \textbf{unilateral} occlusion mask, which corresponds to the areas on left view that are not visible in the right view due to occlusion. It can be computed as:
\begin{align}
\label{mirror_mask}
\mathbf{M}_{\text{RL}}^{\text{L}} &= \min(\sum_{i=0}^{N-1}\mathcal{W}_{\text{R}\rightarrow \text{L}}(\hat{\pi}_{i}^{\text{R}}, d_i), 1) \\
\label{mirror_pi}
\hat{\pi}^{\text{R}}&=\mathcal{S}([\mathcal{W}_{\text{L}\rightarrow \text{R}}(l_i^L, d_i)]_{i=0}^{N-1})
\end{align}
where $\mathcal{W}(\cdot, d)$ is a warping using disparity $d$ and $l_i$ is the logit of the $i$-th plane. L and R represent left and right, respectively.
Taking advantage of the property that the network only predicts artifacts on the left side of objects, they flip the left image and flip the predicted disparity map back, getting a disparity map with only artifacts on the right side of the objects $d_{\text{ff}}$, where the disparity to the left of the objects is accurate. Thus they can supervise artifacts areas on the left side of the object through the unilateral mask $\mathbf{M}_{RL}^L$ and the accurate disparity on the left side of the object $d_{\text{ff}}$.

Post-processing is a method commonly used in depth estimation to improve the final prediction~\cite{godard2017unsupervised}, which inputs both the original image and its flipped image to a depth network and then averages the prediction as
$
    d_{\text{pp}} = \frac{1}{2}(d + d_{\text{ff}})
$, 
where $d$ and $d_{\text{ff}}$ are the predicted disparity map of the original image and its flipped image, respectively.

A natural idea is to use $d_{\text{pp}}$ to further improve the accuracy of the self-distillation label. However, as aforementioned, the left-side artifact of $d$ and the right-side artifact of $d_{\text{ff}}$ cause artifacts to exist on both sides of the objects in $d_{\text{pp}}$. In this paper, rather than using unilateral occlusion masks, we leverage the \textbf{bilateral} occlusion masks to filter both sides artifacts in $d_{\text{pp}}$.
By assuming the left view to be the right view, the occlusion mask on the right side of objects $\mathbf{M}_{\text{LR}}^{\text{L}}$ can be obtained simply by swapping $\mathcal{W}_{\text{L}\rightarrow \text{R}}$ and $\mathcal{W}_{\text{R}\rightarrow \text{L}}$ in \cref{mirror_mask} and \cref{mirror_pi}.

In this way, we can obtain occlusion-aware self-distillation supervision combined with post-processing as:
\begin{equation}
\label{disp_pp}
d_{\text{sd}} = \mathbf{M}_{\text{RL}}^{\text{L}}(\mathbf{M}_{\text{LR}}^{\text{L}}d_{\text{pp}}+(1-\mathbf{M}_{\text{LR}}^{\text{L}})d)+(1-\mathbf{M}_{\text{RL}}^{\text{L}})d_{\text{ff}}
\end{equation}
where $d_{ff}$ is the flipped disparity map of the flipped input, as shown in ~\cref{fig:self_distillation}. We use $L_1$ loss on the predicted disparity in the self-distillation stage as
$
\mathcal{L}_{\text{sd}}=||\hat{d}-d_{\text{sd}}||_1
$.

We also modify our mixture Laplace and perceptual loss in the self-distillation stage using the occlusion mask in the right view as in \cite{gonzalezbello2020forget}. We get the right view occlusion mask $\mathbf{M}^{\text{R}}$ by using \cref{mirror_mask} to warp the weight $\pi$ from the left view to the right. Thus we use $\mathbf{M}^{\text{R}}$ to mask out the loss of the occluded areas in the right view, getting $\mathcal{L}^M_{\text{MLL}}$ and $\mathcal{L}^M_{\text{pc}}$\footnote{More details are provided in the supplementary material.}.


Therefore, our final loss in the self-distillation stage is:
\begin{equation}
\mathcal{L}_{\text{distill}}=\mathcal{L}^M_{\text{MLL}} + \lambda_{1}\mathcal{L}^M_{\text{pc}} + \lambda_{2}\mathcal{L}_{\text{ds}} + \lambda_{3}\mathcal{L}_{\text{sd}}
\end{equation}
which is averaged over pixel, view and batch. $\lambda_{1}$, $\lambda_{2}$ and $\lambda_{3}$ are the weight hyper-parameters. 

\section{Expariments}


\subsection{Training Strategy}
We first train our model with $\mathcal{L}$ for 50 epochs with a batch size of 8, where half of input images are obtained by flipping the other half. Then we fine-tune the network with another epoch at high resolution without resizing and cropping data augmentation, which allows the model to utilize two monocular depth cues simultaneously. We then train another 10 epochs with the self-distillation $\mathcal{L}_{\text{distill}}$ with a batch size of 4 still without resizing and cropping at high resolution. More implementation details are provided in the supplementary material.


\subsection{Dataset}
Following previous work~\cite{xu2021digging, peng2021excavating, watson2019self, zhou2022learning, shu2020feature}, we mainly conducted our experiments on the widely used KITTI\cite{geiger2012we} dataset, which provides sequential stereo images and sparse point clouds using stereo cameras and LiDAR mounted on a moving car. We followed the train test splits used in \cite{godard2019digging}. During stereo training, we adopt the Eigen training split\cite{eigen2014depth}, which contains 22,600 training and 888 validation stereo pairs. As for monocular training, we use the Zhou's split\cite{zhou2017unsupervised}, which removes static frames and contains 19,905 training and 2,212 validation stereo pairs. We evaluate our models on the 697 Eigen raw test images\cite{eigen2014depth} and 652 Eigen improved test images\cite{uhrig2017sparsity} using the metrics proposed in~\cite{geiger2012we}, where Eigen improved test set\cite{uhrig2017sparsity} is obtained by warping the ground truth depth of 11 adjacent frames into one frame and filtering by SGM~\cite{hirschmuller2005accurate}. During both Eigen evaluations, all ground truth and predictions are clipped to within 80m and cropped by as in Eigen crop, and evaluated using the metrics proposed in~\cite{geiger2012we}.


\subsection{Performance Comparisons}

\begin{figure*}[!t]
\centering
\includegraphics[width=\linewidth]{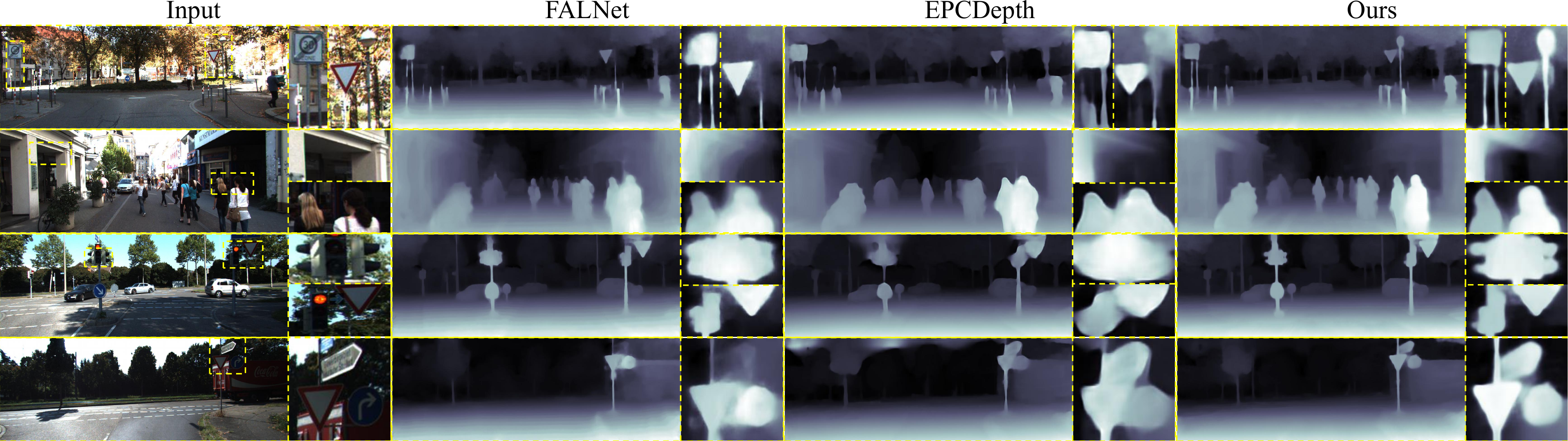}
\caption{\textbf{Qualitative results on the KITTI dataset.} Our network predicts smooth depth for the ground while preserving thin structures and sharp object edges with fewer depth artifacts.}
\label{fig:final_compare}
\end{figure*}

We compare our PlaneDepth against existing state-of-the-art MDE methods on the raw and improved KITTI Eigen test set~\cite{eigen2014depth}, and the quantitative results are shown in \cref{benchmark}. In stereo setting, our method outperforms all competing methods on the Improved Eigen test set~\cite{uhrig2017sparsity} even without using post-processing~\cite{godard2017unsupervised}, which is two times slower because the network needs to make an additional forward for the flipped images. Further, our method performs similarly with or without post-processing thanks to our augmented self-distillation loss. It is worth noting that our proposed augmented self-distillation label further improves performance, but it is slightly less efficient compared to post-processing. Our method also demonstrated consistent improvements when using monocular videos and stereo pairs for training (MS in the table). We also show qualitative comparison against previous state-of-the-art methods in \cref{fig:final_compare} and \cref{fig:top}. Our method predicts continuous depth in the ground region while preserving sharp object edges, recovering more fine details and having fewer depth artifacts around the object boundary compared to other methods. It is also interesting that all methods, including ours, perform slightly worse in the MS setting than in the stereo setting. We attribute it to dynamic objects and inaccurate pose predictions in monocular videos.

\subsection{Ablation Studies}
We have shown improved performance compared to existing methods, now we conduct experiments to validate the effectiveness of each component of our method.

\begin{figure}[!t]
\centering
\includegraphics[width=0.9\linewidth]{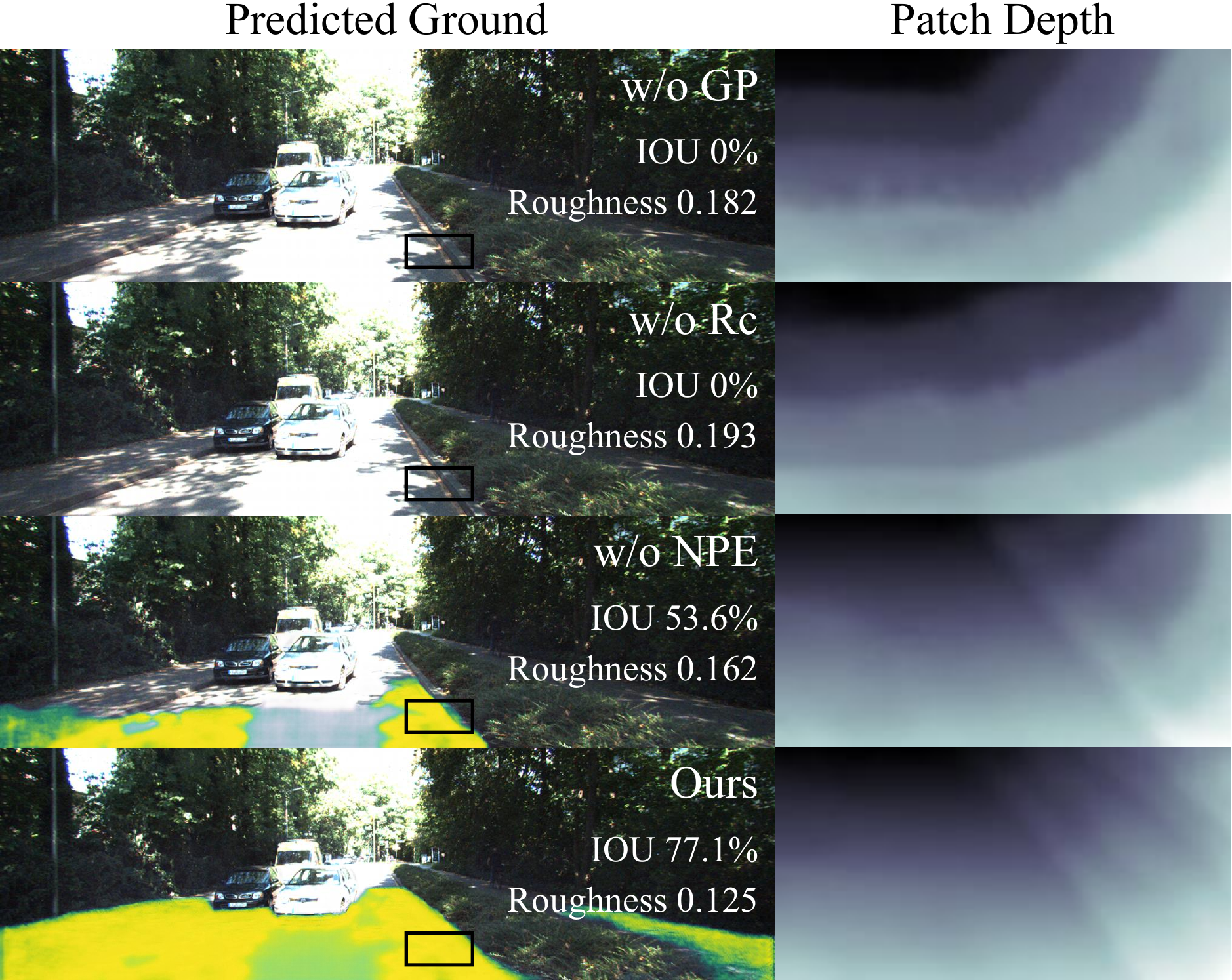}
\caption{\textbf{Predicted Ground}. We use IOU and L2 gradient of disparity map to evaluate the segmentation and roughness of ground on the Eigen test set, respectively. With the help of $\mathbf{R_C}$ and NPE, our method can predict more accurate segmentation and smoother depth for the ground.}
\vspace{0.05cm}
\label{fig:ground_mask}
\end{figure}

\boldparagraph{Orthogonal Planes Prediction}. We have tried various combinations of plane predictions and show the results in \Cref{plane_combinations}. While the quantitative results changed slightly when adding more vertical planes or ground planes,
using ground planes brings additional benefits that we can extract the ground plane without supervision and predict continuous depth as shown in~\cref{fig:ground_mask} and~\cref{fig:top}, respectively, which is important for autonomous driving.
\begin{table}[ht]
\centering
\vspace{-0.1cm}
\resizebox{0.85\linewidth}{!}{
\begin{tabular}{cccccc}
\toprule
\#VP & \#GP & Abs Rel↓ & Sq Rel↓ & Rmse↓ & A1↑   \\ \hline
49 &  0  & 0.090 & \textbf{0.590}    & 4.147             & 0.899 \\
63 &  0  & 0.090 & 0.615             & 4.166             & \textbf{0.900} \\
49 & 14    & \textbf{0.089}    & 0.598             & 4.175             & \textbf{0.900} \\ \hline
\multicolumn{2}{c}{3-axis planes} & 0.090 & 0.598 & \textbf{4.109} & \textbf{0.900} \\ \bottomrule
\end{tabular}
}
\caption{Comparison of different plane combinations, where \#VP and \#GP are the number of vertical and ground planes, respectively. We also evaluate a model using planes with 3-axis normals.}
\vspace{0.1cm}
\label{plane_combinations}
\end{table}

\begin{table*}[!t]
\centering
\resizebox{0.95\linewidth}{!}{
\begin{tabular}{cccccccccccc}
\toprule
Methods     & PP & Network     & Resolution & Train & Abs Rel↓    & Sq Rel↓ & RMSE↓ & RMSE log↓ & A1↑   & A2↑   & A3↑   \\ 
\rowcolor{lightgray} \multicolumn{12}{c}{Raw Eigen Test Set\cite{eigen2014depth}} \\
Monodepth2\cite{godard2019digging}  & \checkmark  & ResNet-50   & 1024×320   & S     & 0.097       & 0.793   & 4.533 & 0.181     & 0.896 & 0.965 & 0.983 \\
DepthHint\cite{watson2019self}   & \checkmark  & ResNet-50   & 1024×320   & S     & 0.096       & 0.710   & 4.393 & 0.185     & 0.890 & 0.962 & 0.981 \\
EPCDepth\cite{peng2021excavating}    & \checkmark  & ResNet-50   & 1024×320   & S     & 0.091       & 0.646   & 4.207 & 0.176     & 0.901 & 0.966 & 0.983 \\
OCFDNet\cite{zhou2022learning} & \checkmark  & ResNet-50-A & 1280×384   & S     & 0.090       & 0.564   & 4.007 & 0.172     & 0.903 & \underline{0.967} & \underline{0.984} \\
FALNet\cite{gonzalezbello2020forget}     &    & FALNet      & 1280×384   & S     & 0.097       & 0.590   & 3.991 & 0.177     & 0.893 & 0.966 & \underline{0.984} \\
FALNet\cite{gonzalezbello2020forget}     & \checkmark  & FALNet      & 1280×384   & S     & 0.093       & 0.564   & \textbf{3.973} & 0.174     & 0.898 & \underline{0.967} & \textbf{0.985} \\
PLADENet\cite{gonzalez2021plade}   &    & PLADENet    & 1280×384   & S     & 0.092       & 0.626   & 4.046 & 0.175     & 0.896 & 0.965 & \underline{0.984} \\
PLADENet\cite{gonzalez2021plade}   & \checkmark  & PLADENet    & 1280×384   & S     & 0.089       & 0.590   & 4.008 & 0.172     & 0.900 & \underline{0.967} & \textbf{0.985} \\ 
\hline
Ours        &    & ResNet-50-A & 1280×384   & S     & \underline{0.085}       & \underline{0.563}   & 4.023 & \underline{0.171}     & \underline{0.910} & \textbf{0.968} & \underline{0.984} \\
Ours        & \checkmark  & ResNet-50-A & 1280×384   & S     & \textbf{0.084} & \textbf{0.549}   & \underline{3.981} & \textbf{0.169}     & \textbf{0.911} & \textbf{0.968} & \underline{0.984} \\ 
Ours\dag     & \checkmark  & ResNet-50-A & 1280×384   & S     & 0.083 & 0.533   & 3.919 & 0.167     & 0.913 & 0.969 & 0.985 \\ 
\hline
\hline
Monodepth2\cite{godard2019digging}  & \checkmark  & ResNet-18   & 1024×320   & MS    & 0.104       & 0.775   & 4.562 & 0.191     & 0.878 & 0.959 & 0.981 \\
DepthHint\cite{watson2019self}   & \checkmark  & ResNet-18   & 1024×320   & MS    & 0.098       & 0.702   & 4.398 & \underline{0.183}     & 0.887 & \textbf{0.963} & \textbf{0.983} \\
FeatureNet\cite{shu2020feature}  &    & ResNet-50   & 1024×320   & MS    & 0.099       & 0.697   & 4.427 & 0.184     & 0.889 & \textbf{0.963} & \underline{0.982} \\ \hline
Ours* &    & ResNet-50-A & 1280×384   & MS    & \underline{0.092}       & \underline{0.601}   & \underline{4.188} & 0.184     & \underline{0.893} & 0.961 & 0.981 \\ 
Ours* & \checkmark  & ResNet-50-A & 1280×384   & MS    & \textbf{0.090}       & \textbf{0.584}   & \textbf{4.130} & \textbf{0.182}     & \textbf{0.896} & \underline{0.962} & 0.981 \\
\rowcolor{lightgray} \multicolumn{12}{c}{Improved Eigen Test Set\cite{uhrig2017sparsity}} \\
Monodepth2\cite{godard2019digging}  & \checkmark  & ResNet-50   & 1024×320   & S     & 0.077       & 0.455   & 3.489 & 0.119     & 0.933 & 0.988 & 0.996 \\
DepthHint\cite{watson2019self}   & \checkmark  & ResNet-50   & 1024×320   & S     & 0.074       & 0.364   & 3.202 & 0.114     & 0.936 & 0.989 & 0.997 \\
OCFDNet\cite{zhou2022learning}     & \checkmark & ResNet-50-A & 1280×384   & S     & 0.070       & 0.262   & 2.786 & 0.103     & 0.951 & 0.993 & \underline{0.998} \\
FALNet\cite{gonzalezbello2020forget}      & \checkmark & FALNet & 1280×384   & S     & 0.071       & 0.281   & 2.912 & 0.108     & 0.943 & 0.991 & \underline{0.998} \\
PLADENet\cite{gonzalez2021plade}   & \checkmark  & PLADENet & 1280×384   & S     & \underline{0.066}       & 0.272   & 2.918 & 0.104     & 0.945 & 0.992 & \underline{0.998} \\ %
\hline
Ours     & & ResNet-50-A & 1280×384   & S     & \textbf{0.063}       & \underline{0.245}   & 2.718 & \underline{0.096}    & \underline{0.959} & \underline{0.994} & \underline{0.998} \\
Ours     & \checkmark & ResNet-50-A & 1280×384   & S     & \textbf{0.063}  & \textbf{0.236}   & \textbf{2.674} & \textbf{0.095}     & \textbf{0.960} & \textbf{0.994} & \textbf{0.999} \\ 
Ours\dag & \checkmark & ResNet-50-A & 1280×384   & S     & 0.062  & 0.227   & 2.609 & 0.093     & 0.961 & 0.995 & 0.999 \\ 
\bottomrule
\end{tabular}
}
\caption{Comparison of performance on KITTI Eigen test set\cite{eigen2014depth}. The best is in \textbf{bold} and the second best is \underline{underlined} in each metric. S stands for stereo training using left and right views and a fixed baseline and MS stands for both stereo and monocular training adding front and rear frames of the left view without camera poses. ResNet-50-A represents U-Net\cite{ronneberger2015u} with DenseASPP module\cite{yang2018denseaspp} using ResNet-50\cite{he2016deep} as the backbone. PP with checkmark\checkmark refers to post-processing\cite{zhou2017unsupervised}. *: Due to the limitation of memory usage for high-resolution training, our MS training is only performed at low-resolution in stage one without self-distillation. \dag: Post-processing with our method for generating self-distillation labels.}
\label{benchmark}
\end{table*}

\boldparagraph{Mixture Laplace Loss}. Quantitative results in~\Cref{loss_comparison} show that our mixture Laplace loss outperforms simple L1 loss used in baseline methods~\cite{gonzalezbello2020forget, gonzalez2021plade}.
To further quantify the effects of different losses, we propose a new metric: Mean Maximum Probability (MMP)\footnote{Definition is provided in the supplementary material.} which reflects the confidence of the network's prediction on certain planes. As shown in~\Cref{loss_comparison}, using MML results in higher accuracy and MMP, indicating that MML is more deterministic and less ambiguous compared to using L1 loss on the composite image.



\begin{table}[t]
\centering
\resizebox{1\linewidth}{!}{
\begin{tabular}{ccccccc}
\toprule
Loss & Plane & Abs Rel↓ & Sq Rel↓ & Rmse↓ & A1↑   & MMP   \\ \hline
L1  & VP & 0.093          & 0.624            & 4.402          & 0.891 & 0.574 \\
L1  & OP & 0.095          & 0.606            & 4.293          & 0.893 & 0.543 \\
MLL & VP & 0.090          & \textbf{0.590}   & \textbf{4.147} & 0.899 & 0.610 \\
MLL & OP & \textbf{0.089} & 0.598            & 4.175          & \textbf{0.900} & 0.607 \\ \bottomrule
\end{tabular}
}
\caption{Comparison of different photometric loss, where MLL denotes our mixture Laplace loss. MMP stands for Mean Maximum Probability. VP means only using the vertical planes, OP means using our orthogonal planes. The result shows that MLL is effective for plane-based methods and makes predictions more focused on a specific plane.}
\label{loss_comparison}
\end{table}

\boldparagraph{Orthogonality-preserved Data Augmentation}. We show the quantitative ablation of transformation matrix $\mathbf{R_C}$ and neural positional encoding (NPE) in~\cref{tab:npe_rc} and qualitative comparison on ground prediction in~\cref{fig:ground_mask}. Without using $\mathbf{R_C}$ to rectify predefined ground planes, the tilt of the ground makes the network relies on vertical planes completely, 
resulting in discontinuity in the ground region. 
Without NPE, the network has difficulty in predicting the tilt of the ground plane, resulting in discontinuous depth patches. In contrast, with the help of $\mathbf{R_C}$ and NPE, our method can utilize ground planes easily, resulting in more accurate ground segmentation and smoother ground depth. 

\begin{table}[t]
\centering
\resizebox{0.85\linewidth}{!}{
\begin{tabular}{cccccc}
\toprule
 $\mathbf{R_C}$ & NPE          & Abs Rel↓        & Sq Rel↓           & Rmse↓ & A1↑   \\ \hline
                &              & 0.092           & 0.596   &  4.143    & 0.898   \\
                & \checkmark   & 0.090           &  \textbf{0.593}   & 4.186     & 0.899 \\
\checkmark      &              & 0.094           & \textbf{0.593} & \textbf{4.141}    & 0.898 \\
\checkmark      & \checkmark   & \textbf{0.089}  & 0.598             & 4.175             & \textbf{0.900} \\ 
 \bottomrule
\end{tabular}
}
\caption{Comparison of ground prediction. Smoother ground depth with the help of $\mathbf{R_C}$ and NPE brings a performance boost.}
\label{tab:npe_rc}
\end{table}

\boldparagraph{Augmented Self-distillation}. We first compare the accuracy of pseudo labels generated by different methods. As shown in~\cref{label_comparison}, our proposed bilateral mask combined with post-processing generates more accurate labels than other methods, which is beneficial for self-distillation. Further, we compare quantitative results with or without self-distillation loss in \cref{distillation_stage}, which verifies that the performance improvement is due to our self-distillation loss rather than the longer training time. 

\begin{table}[t]
\centering
\resizebox{0.8\linewidth}{!}{
\begin{tabular}{ccccc}
\toprule
Strategy & Abs Rel↓ & Sq Rel↓ & Rmse↓ & A1↑    \\ \hline
LR   & 0.090    & 0.616   & 4.188 & 0.898  \\
HR  & 0.090    & 0.630   & 4.270 & 0.898  \\
Ours  & \textbf{0.086}    & \textbf{0.581}   & \textbf{4.094} & \textbf{0.906}  \\
\bottomrule
\end{tabular}
}
\caption{Comparison of different training strategies. LR and HR denote low resolution ($640\times 192$) and high resolution ($1280\times 384$) training, respectively. The results show that our approach of combining LR training with one additional HR epoch, utilizing both monocular depth cues, significantly improvements performance.}
\vspace{0.05cm}
\label{label_strategy}
\end{table}

\begin{table}[t]
\centering
\resizebox{0.8\linewidth}{!}{
\begin{tabular}{ccccc}
\toprule
Source & Abs Rel↓ & Sq Rel↓ & Rmse↓ & A1↑    \\ \hline
PP   & 0.085    & 0.557   & 4.020 & 0.909  \\
UM  & 0.084    & 0.552   & 3.982 & 0.912  \\
PP+BM  & \textbf{0.083}    & \textbf{0.540}   & \textbf{3.940} & \textbf{0.913}  \\
\bottomrule
\end{tabular}
}
\caption{Comparison of different distillation label, where PP denotes post-processing~\cite{godard2017unsupervised}, UM denotes unilateral occlusion mask~\cite{gonzalezbello2020forget}, and PP+BM denotes our bilateral mask combined with post-processing. Our proposed method generate more accurate self-distillation labels.}
\label{label_comparison}
\end{table}

\begin{table}[!t]
\centering
\resizebox{0.8\linewidth}{!}{
\begin{tabular}{ccccc}
\toprule
Loss & Abs Rel↓ & Sq Rel↓ & Rmse↓ & A1↑    \\ \hline
$\mathcal{L}$ & 0.086    & 0.584   & 4.132 & 0.905  \\
$\mathcal{L}_{\text{distill}}$ & \textbf{0.085}    & \textbf{0.563}   & \textbf{4.023} & \textbf{0.910}  \\
\bottomrule
\end{tabular}
}
\caption{Ablation study on self-distillation. Result shows that using self-distillation loss $\mathcal{L}_{\text{distill}}$ significantly improves performance compared to naively increasing training time.}
\label{distillation_stage}
\end{table}

\section{Conclusion}
We propose PlaneDepth, a novel orthogonal planes based depth representation for self-supervised depth estimation, which enables segmenting and predicting a continuous ground plane compared to existing frontal-parallel planes based representations. The depth of the scene is represented as a mixture of Laplacian of the orthogonal planes. Considering that traditional data augmentation may break the orthogonality constraint of the planes, we solve this explicitly by calculating the resizing cropping transformation as well as the neural position encoding. We further propose to improve the final prediction by incorporating post-processing into the self-distillation objective via our bilateral occlusion masks. Extensive experiments on KITTI validate the effectiveness and efficiency of our proposed approach.

\boldparagraph{Limitations}. We do not constrain the choice of different types of planes and the inclination caused by car bumping, so our unsupervised ground segmentation may not be robust enough to be directly used in safety-critical applications. Based on this, combining with other unsupervised ground segmentation methods, or explicitly modeling car tilt may be interesting future directions.

\boldparagraph{Acknowledgements}. This work was supported by National Key R\&D Program of China (2018AAA0100704), NSFC \#61932020, \#62172279, Science and Technology Commission of Shanghai Municipality (Grant No.20ZR1436000), Program of Shanghai Academic Research Leader, and "Shuguang Program" supported by Shanghai Education Development Foundation and Shanghai Municipal Education Commission.


\appendix

\setcounter{page}{1}

\twocolumn[
\centering
\Large
\textbf{PlaneDepth: Self-supervised Depth Estimation via Orthogonal Planes} \\
\vspace{0.5em}Supplementary Material \\
\vspace{1.0em}
] 

\section{Introduction}
We first provide additional details of our method in \cref{sm_sec:mthod}, specifically regarding the resizing cropping transformation and the self-distillation loss. Furthermore, we provide implementation details in \cref{sm_sec:implement}. Additionally, we report more experimental results and examine the influence of different training strategies in \cref{sm_sec:experiment}.

\section{Additional Method Details}
\label{sm_sec:mthod}
\subsection{Resizing Cropping Transformation}
Resizing and cropping data augmentation is expected to only modify the camera intrinsics while keeping the world coordinates unchanged. However, the networks must predict the depth and relative pose of the original images for all augmented inputs, which results in the disruption of a crucial monocular cue - the closer an object is, the lager its relative size~\cite{bello2022positional}. We therefore assume that all input images are captured by the same camera system, and discuss the effect of the resizing cropping augmentation on the world coordinates.

\begin{figure}[ht]
\centering
\includegraphics[width=\linewidth]{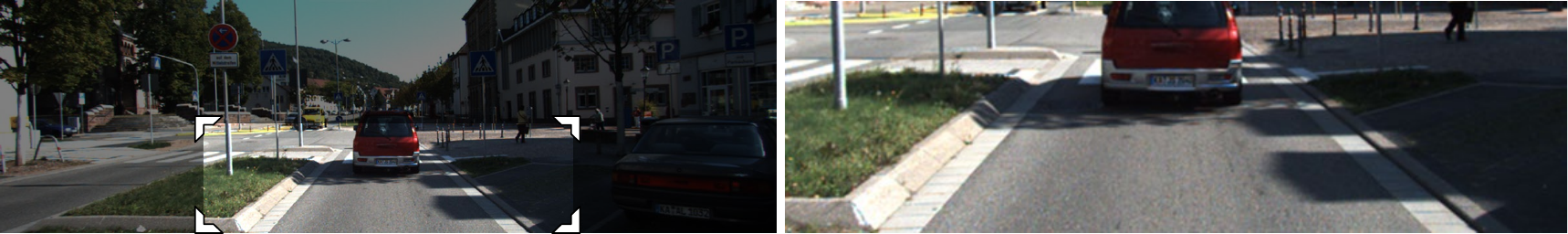}
\caption{Image is cropped at coordinates $(p_x, p_y)$ using a scale factor of $f_s$, and then resized to the original resolution.}
\label{sm_fig:resizing_cropping}
\end{figure}

Given the center pixel coordinates $\mathbf{P}_c=(p_x, p_y)$ and the cropping scale factor $f_s$, the pixel coordinates $\mathbf{P}=(x,y)$ and depth $D$ of the original image after resizing and cropping are modified as follows:
\begin{equation}
\widetilde{\mathbf{P}} = \frac{\mathbf{P}-\mathbf{P}_c}{f_s}+\frac{\mathbf{S}}{2} \hspace{1cm} \widetilde{D}=f_s D,
\end{equation}
where $\mathbf{S}=(W,H)$ is the size of the image. The adjustment in depth is based on the assumption that when the relative size of an object increases by a factor of $f_s$, its depth decreases by the same factor, $f_s$.

Hence, the rectified coordinates $\tilde{\mathbf{w}}$ obtained from augmented images and the original world coordinates $w$ can be expressed as:
\begin{equation}
\label{sm_eq:w_wn}
\tilde{\mathbf{w}} = \widetilde{D}\mathbf{K}^{-1}\begin{bmatrix}
\tilde{x} \\ 
\tilde{y} \\
1
\end{bmatrix}
\hspace{1cm}
\mathbf{w} = D\mathbf{K}^{-1}\begin{bmatrix}
x \\ 
y \\
1
\end{bmatrix}.
\end{equation}

The transformation $\mathbf{R_C}$ of world coordinates from $\mathbf{w}$ to $\tilde{\mathbf{w}}$ is given by:
\begin{equation}
\label{sm_eq:rc}
\tilde{\mathbf{w}} = \mathbf{R_C}\mathbf{w} \hspace{1cm} \mathbf{R_C}=
\begin{bmatrix}
1 &   & \frac{c_x-p_x}{f_x} \\ 
  & 1 & \frac{c_y-p_y}{f_y} \\
  &   & f_s
\end{bmatrix},
\end{equation}
where $(f_x,f_y)$ represents the focal length and $(c_x,c_y)$ denotes the principal point of the camera.

Given the equations of the transformed plane and the original plane:
\begin{equation}
    \tilde{\mathbf{n}}^T\mathbf{R_C}\mathbf{w}-\tilde{\delta}=0 \hspace{1cm} \mathbf{n}^T\mathbf{w}-\delta=0,
\end{equation}
the rectified normal $\tilde{\mathbf{n}}$ and distance $\tilde{\delta}$ can be computed as:
\begin{equation}
\label{sm_eq:planes_rc}
\tilde{\mathbf{n}} = \frac{\mathbf{R_C}^{-T}}{||\mathbf{R_C}^{-T}\mathbf{n}||}\mathbf{n}\hspace{1cm} \tilde{\delta} = \frac{\delta}{||\mathbf{R_C}^{-T}\mathbf{n}||}.
\end{equation}
Therefore, given the original vertical plane normal $\mathbf{n}^v=[0\ 0\ 1]^T$ and ground plane normal $\mathbf{n}^g=[0\ 1\ 0]^T$, the rectified vertical plane normal $\tilde{\mathbf{n}}^v$ and ground plane normal $\tilde{\mathbf{n}}^g$ are:
\begin{align}
\label{sm_eq:normals_rc}
\tilde{\mathbf{n}}^v = \frac{\mathbf{R_C}^{-T}\mathbf{n}^v}{||\mathbf{R_C}^{-T}\mathbf{n}||} &= \begin{bmatrix}
0 \\ 
0 \\
1
\end{bmatrix}
\\
\tilde{\mathbf{n}}^g = \frac{\mathbf{R_C}^{-T}\mathbf{n}^g}{||\mathbf{R_C}^{-T}\mathbf{n}||} &= \frac{1}{\sqrt{1+(\frac{p_y-c_y}{f_yf_s})^2}}\begin{bmatrix}
0 \\ 
1 \\
\frac{p_y-c_y}{f_yf_s}
\end{bmatrix}.
\end{align}
Consequently, the normal of the planes $\tilde{\mathbf{n}}^v$ and $\tilde{\mathbf{n}}^g$ after transformation are no longer orthogonal.

Furthermore, two views which are resized and cropped by the same parameters have the relationship as:
\begin{equation}
\mathbf{R_C}\mathbf{w}^r = 
\widetilde{\mathbf{R}}
\mathbf{R_C}\mathbf{w} + \widetilde{\mathbf{t}}
\hspace{1cm}
\mathbf{w}^r = 
\mathbf{R}\mathbf{w} + \mathbf{t},
\end{equation}
where $\mathbf{R}$ and $\mathbf{t}$ denote the rotation and translation, respectively. Therefore, the modified relative pose can be expressed as:
\begin{equation}
\widetilde{\mathbf{R}}=\mathbf{R_C}\mathbf{R}\mathbf{R_C}^{-1}\hspace{1cm} \tilde{\mathbf{t}}=\mathbf{R_C}\mathbf{t}
\end{equation}

\subsection{Self-distillation Loss}
Taking only the left view as input, we now provide the details of filtering the loss of the occluded pixels in the right view in self-distillation strategy.

Similar to \cite{gonzalezbello2020forget}, we obtain the right view occlusion mask $\mathbf{M}^\text{R}$ by using:
\begin{equation}
\label{sm_eq:target_mask}
\mathbf{M}^\text{R} = \min(\sum_{i=0}^{N-1}\mathcal{W}_{\text{L}\rightarrow \text{R}}(\pi_{i}, d_i), 1),
\end{equation}
where $\mathcal{W}(\cdot, d)$ represents the warping function using disparity $d$, and $\pi_{i}$ denotes the predicted Laplace weight of the $i$-th plane. We define the occlusion-aware mixture Laplace loss and perceptual loss as:
\begin{align}
&\mathcal{L}^\text{M}_{\text{MLL}} = \mathbf{M}^\text{R} \odot (-\log\sum_{i=0}^{N-1} \frac{\hat{\pi}_{i}e^{\frac{-\frac{1}{3}||\mathbf{I}_r - \hat{\mathbf{I}}_{i}||_1}{\hat{\sigma}_{i}}}}{2\hat{\sigma}_{i}}) \\
&\mathcal{L}^\text{M}_{\text{pc}} = ||\phi_l(\mathbf{I}_r)-\phi_l(\mathbf{M}^R \odot \hat{\mathbf{I}}_r + (1-\mathbf{M}^R)\odot \mathbf{I}_r)||^2_2,
\end{align}
where $\hat{\pi}_i,\ \hat{\sigma}_i,\ \hat{\mathbf{I}}_i$ are the weight, scale and image warped by the $i$-th plane, respectively. $\mathbf{I}_r$ is the right view image, $\hat{\mathbf{I}}_r$ is the synthesised right view obtained by compositing $\hat{\mathbf{I}}_i$, and $\phi_l$ is the first $l$ maxpool layers of a VGG19~\cite{simonyan2014very} model pre-trained on the ImageNet~\cite{deng2009imagenet}.

Therefore, our final loss in the self-distillation stage is:
\begin{equation}
\mathcal{L}_{\text{distill}}=\mathcal{L}^M_{\text{MLL}} + \lambda_{1}\mathcal{L}^M_{\text{pc}} + \lambda_{2}\mathcal{L}_{\text{ds}} + \lambda_{3}\mathcal{L}_{\text{sd}}
\end{equation}
which is averaged over pixel, view and batch. $\lambda_{1}$, $\lambda_{2}$ and $\lambda_{3}$ are the weight hyper-parameters. 

\section{Implement Details}
\label{sm_sec:implement}
\subsection{Training Details}
We implement our network using PyTorch~\cite{paszke2019pytorch} and train it using Adam~\cite{kingma2014adam} with $\beta_1=0.5, \beta_2=0.999$. Our default data augmentations consist of resizing using a random scaling factor between 0.75 and 1.5, random cropping using size $640\times 192$, and random gamma, brightness and color augmentations. In the first stage, we train our model with $\mathcal{L}$ for 50 epochs using a batch size of 8, where half of input images are obtained by horizontal flipping the other half. Our initialization learning rate is $1\times 10^{-4}$, which is halved at 30 and 40 epochs, respectively. We then fine-tune the network with another epoch at high resolution $(1280\times 384)$ without resizing and cropping data augmentation. Subsequently, we train another 10 epochs using the self-distillation loss $\mathcal{L}_{\text{distill}}$ with a batch size of 4 at high resolution, still without resizing and cropping. or self-distillation, we use an initialization learning rate of $2\times 10^{-5}$, which is halved at 5 epoch. We set the minimum and maximum disparity to $d_\text{min}=2$ and $d_\text{max}=300$, and the minimum and maximum camera heights to $h_\text{min}=1$ and $h_\text{max}=2$. Additionally, we set the hyper-parameters of our loss function to $\lambda_1=0.1,\ \lambda_2=0.04,\ \lambda_3=1$.

\subsection{Network Architecture}
We adopt the U-Net~\cite{ronneberger2015u} with DenseASPP~\cite{yang2018denseaspp} module as the base network for depth estimation, as proposed by Zhou and Dong~\cite{zhou2022learning}. To improve the accuracy of the ground depth and segmentation, we further enhance our PlaneDepth network by incorporating neural positional encoding (NPE). Similar to \cite{godard2019digging}, we simply use the first five blocks of ResNet50~\cite{he2016deep} as the encoder, and the five upsampling blocks proposed in \cite{godard2019digging} as our decoder. Consistent with \cite{zhou2022learning}, we insert a DenseASPP module~\cite{yang2018denseaspp} with dilation rates of $r\in\{3,6,12,18,24\}$ between the first two blocks of the decoder. To assist the network in learning resizing and cropping information, we encode grid $\mathbf{g}$ as an 8-channel feature map using NPE following \cite{gonzalez2021plade}, and concatenate it with the input of the first four decoder blocks.

For our pose estimation network, following Godard \etal~\cite{godard2019digging}, we use the first five blocks of ResNet18~\cite{he2016deep} with double input channels as the encoder to extract two view features. We then employ three convolution layers followed by global average pooling for the decoder. Additionally, we concatenate an 8-channel resizing and cropping feature with the input of the decoder.

\subsection{Evaluation Metrics}
The metrics used for evaluation are defined as:
\begin{enumerate}
    \item $\text{Abs Rel}=\frac{1}{n}\sum_{i=1}^{n}\frac{|D^*_i-D_i|}{D^*}$
    \item $\text{Sq Rel}=\frac{1}{n}\sum_{i=1}^{n}\frac{(D^*_i-D_i)^2}{D^*}$
    \item $\text{RMSE}=(\frac{1}{n}\sum_{i=1}^{n}(D^*_i-D_i)^2)^{1/2}$
    \item $\text{RMSE log}=(\frac{1}{n}\sum_{i=1}^{n}(\log D^*_i-\log D_i)^2)^{1/2}$
    \item $\text{A}_t=\frac{1}{n}|\{D_i|i\leq n, max(\frac{D_i}{D^*_i},\frac{D^*_i}{D_i}) < 1.25^t\}|$
    \item $\text{MMP}=\frac{1}{n}\sum_{i=1}^{n}(\max\limits_{0\leq j\leq N-1} p_j \big/ \sum_{k=0}^{N-1}p_k)$
\end{enumerate}
where $D^*$ and $D$ represent the ground truth and the predicted depth map, respectively. $p_i$ denotes the probability that the pixel belongs to the $i$-th plane.

\section{Additional Experiments}
\label{sm_sec:experiment}

\begin{table*}[t]
\centering
\resizebox{\textwidth}{!}{
\begin{tabular}{cccccccccccc}
\toprule
Stage1 & Stage1 & Finetune & Finetune & \multicolumn{4}{c}{LR performance} & \multicolumn{4}{c}{HR performance}   \\ 
Resolution & RC & Resolution & RC & Abs Rel↓ & Sq Rel↓ & Rmse↓ & A1↑ & Abs Rel↓ & Sq Rel↓ & Rmse↓ & A1↑ \\ \hline
LR &  & LR &  & \underline{0.103} & 0.748 & \underline{4.631} & \underline{0.878} & 0.461 & 5.342 & 8.675 & 0.414 \\
HR &  & HR &  & 0.174 & 1.284 & 6.215 & 0.713 & \underline{0.090} & 0.630 & 4.270 & \underline{0.898} \\
LR & \checkmark & LR & \checkmark & \textbf{0.100} & \textbf{0.708} & \textbf{4.588} & \textbf{0.882} & \underline{0.090} & \underline{0.616} & \underline{4.188} & \underline{0.898} \\
LR & \checkmark & HR &  & \textbf{0.100} & \underline{0.713} & 4.653 & 0.877 & \textbf{0.086} & \textbf{0.581} & \textbf{4.094} & \textbf{0.906} \\ 
\bottomrule
\end{tabular}
}
\caption{Comparison of different training strategies in the first training stage and the fine-tune epoch. LR denotes low resolution ($640\times 192$), HR represents high resolution ($1280\times 384$), and RC indicates resizing and cropping data augmentation. The results show that exploiting both monocular cues in our training strategy significantly improves performance.}
\label{strategy_comparison}
\end{table*}

\subsection{Ablation of Training Strategy}
Without any special data augmentation, the depth estimation network will predict depth based solely on the vertical image position cue~\cite{dijk2019neural}. Gonzalez and Kim~\cite{gonzalezbello2020forget} showed that Resizing and cropping corrupt the vertical image position of object, enabling the network to exploit the relative object size cue. Furthermore, since disparity and relative object size remain consistent, a network that predicts disparity using the relative size cue can naturally adapt to inputs of various resolutions.

Since the vertical image position cue of outdoor scenes is also closely related to depth, exploiting this cue can further improve the performance. Taking advantage of the fact that networks tend to learn the vertical image position, we propose fine-tuning the network for an additional epoch at high resolution without resizing and cropping at the end of training. This fine-tuning step can further introduce vertical image position cue and adapt the network to real images, resulting a performance boost as shown in \cref{strategy_comparison}.

\Cref{strategy_comparison} shows that networks trained without resizing and cropping are limited to performing only at the resolution of the training data. However, networks trained using resizing and cropping augmentation perform well in both low and high resolutions.
Compared with training on HR all the time, our strategy of training on LR and fine-tuning on HR saves a lot of training time and performs better, indicating that the performance improvement is due to the utilization of both cues rather than the increase in training resolution.
Compared with training on LR with RC all the time, our method learns the vertical image position cue at high resolution during fine-tuning. This improves performance at HR but affects performance at LR since the vertical image position cue is not universal across different resolutions.

\begin{figure}[t]
\centering
\includegraphics[width=\linewidth]{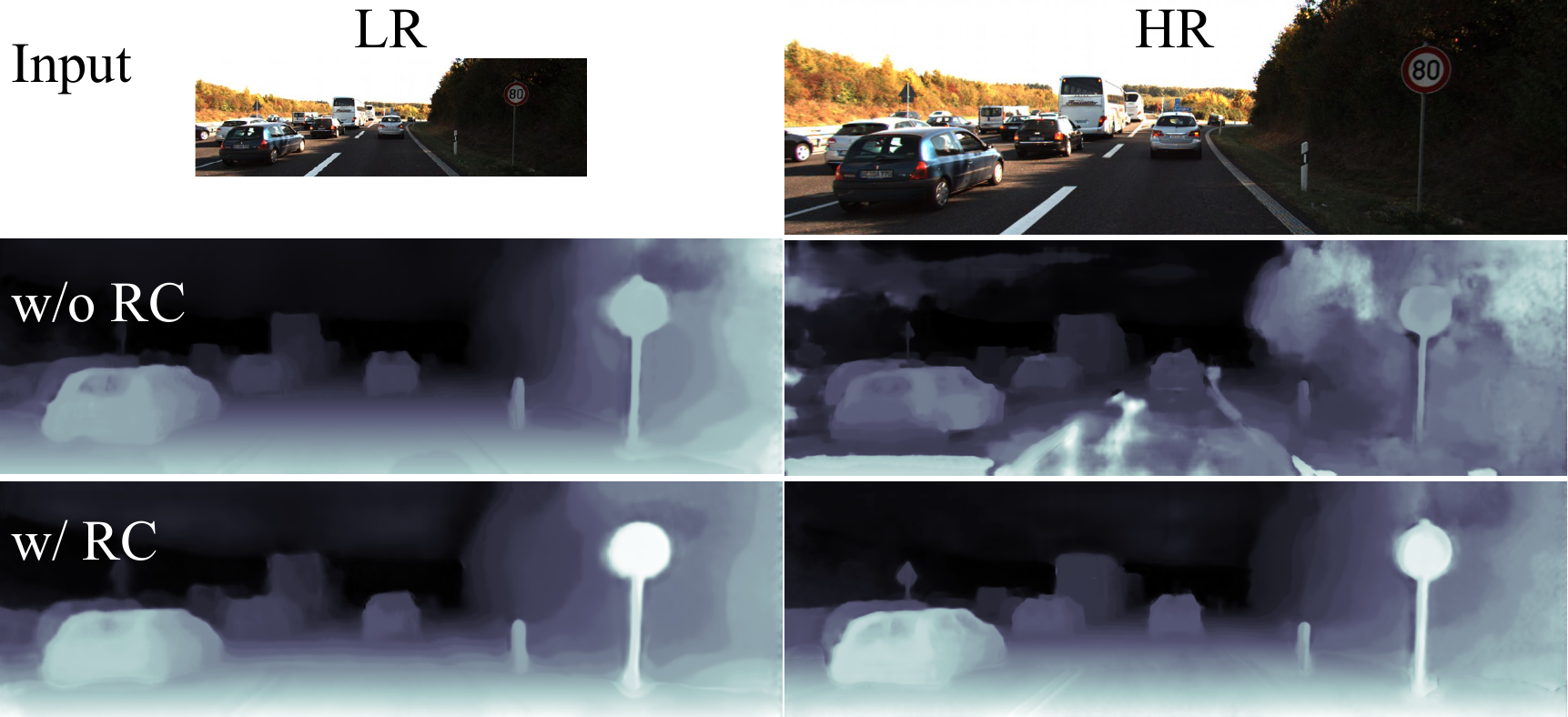}
\caption{\textbf{Visualization of the effect of resizing cropping.} LR and HR are low and high resolution, respectively. RC means training with resizing and cropping. The prediction results of inputs with different resolutions are resized to the same resolution for evaluation. Network trained using resizing and cropping can naturally adjust to inputs with different resolutions.}
\label{fig:train_strategy}
\end{figure}

\begin{figure}[t]
\centering
\includegraphics[width=\linewidth]{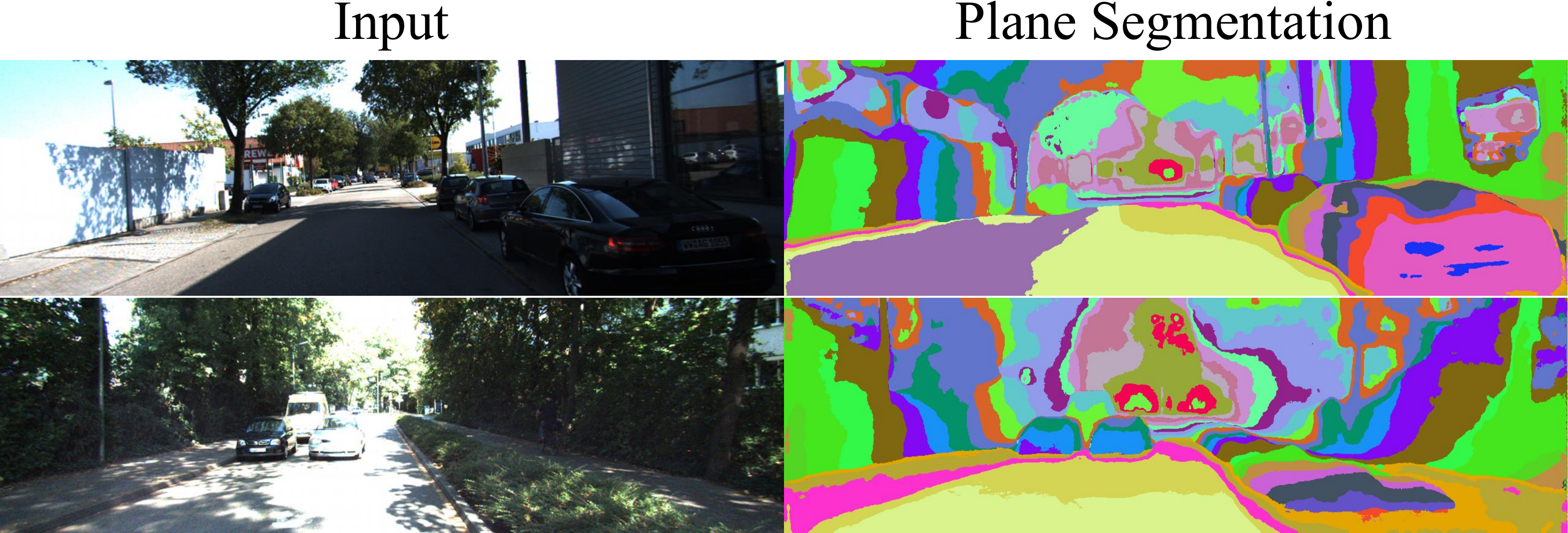}
\caption{\textbf{Visualization of regions modeled by different planes.} The ground is modeled by composing different ground planes, which is necessary since the ground is not always an ideal horizontal plane.}
\label{fig:plane_segmentation}
\end{figure}

\subsection{Monocular Training Results}

\begin{table}[t]
\begin{center}
\resizebox{\linewidth}{!}{
\begin{tabular}{ccccccc}
\toprule
Train & NPE & AM & Abs Rel↓ & Sq Rel↓ & Rmse↓ & A1↑   \\ \hline
S &  &   & 0.089 & 0.598 & 4.175 & 0.900 \\ \hline
MS & \checkmark &   & 0.109 & 1.183 & 5.034 & 0.877 \\
MS &    & \checkmark & \underline{0.093} & \textbf{0.599} & \underline{4.203} & \underline{0.891} \\
MS & \checkmark & \checkmark & \textbf{0.092} & \underline{0.601} & \textbf{4.188} & \textbf{0.893} \\
M & \checkmark & \checkmark   & 0.348 & 3.842 & 11.383 & 0.406 \\
M\dag &  & \checkmark   & 0.113 & 1.049 & 4.943 & 0.859 \\\bottomrule
\end{tabular}
}
\caption{Comparison of different monocular training settings. NPE refers to adding NPE as input to the pose network. AM is automask proposed by \cite{godard2019digging}. S stands for stereo training using left and right views with a fixed baseline. M denotes monocular training using front and rear frames without camera poses of the left view as reference views. MS stands for both stereo and monocular training. All depth networks use NPE inputs. \dag: We use the pretrained pose network provided by \cite{godard2019digging} to solve the scale ambiguity.}
\label{MS_training}
\end{center}
\end{table}

We conduct experiments with various monocular training settings and show results in \cref{MS_training}. We find that adding monocular supervision without considering appropriate measures significantly decreases performance. However, when using automask~\cite{godard2019digging} and NPE, this negative impact can be mitigated. When stereo supervision is not used, our predefined planes are no longer in suitable positions due to the scale ambiguity, leading to training failure. To address this issue, we try to use the pretrained pose network provided by \cite{godard2019digging} for predicting poses with suitable scales and achieve good performance. This result confirms the correctness of our resizing cropping transformation of camera pose.

\begin{figure*}[!t]
\centering
\includegraphics[width=\linewidth]{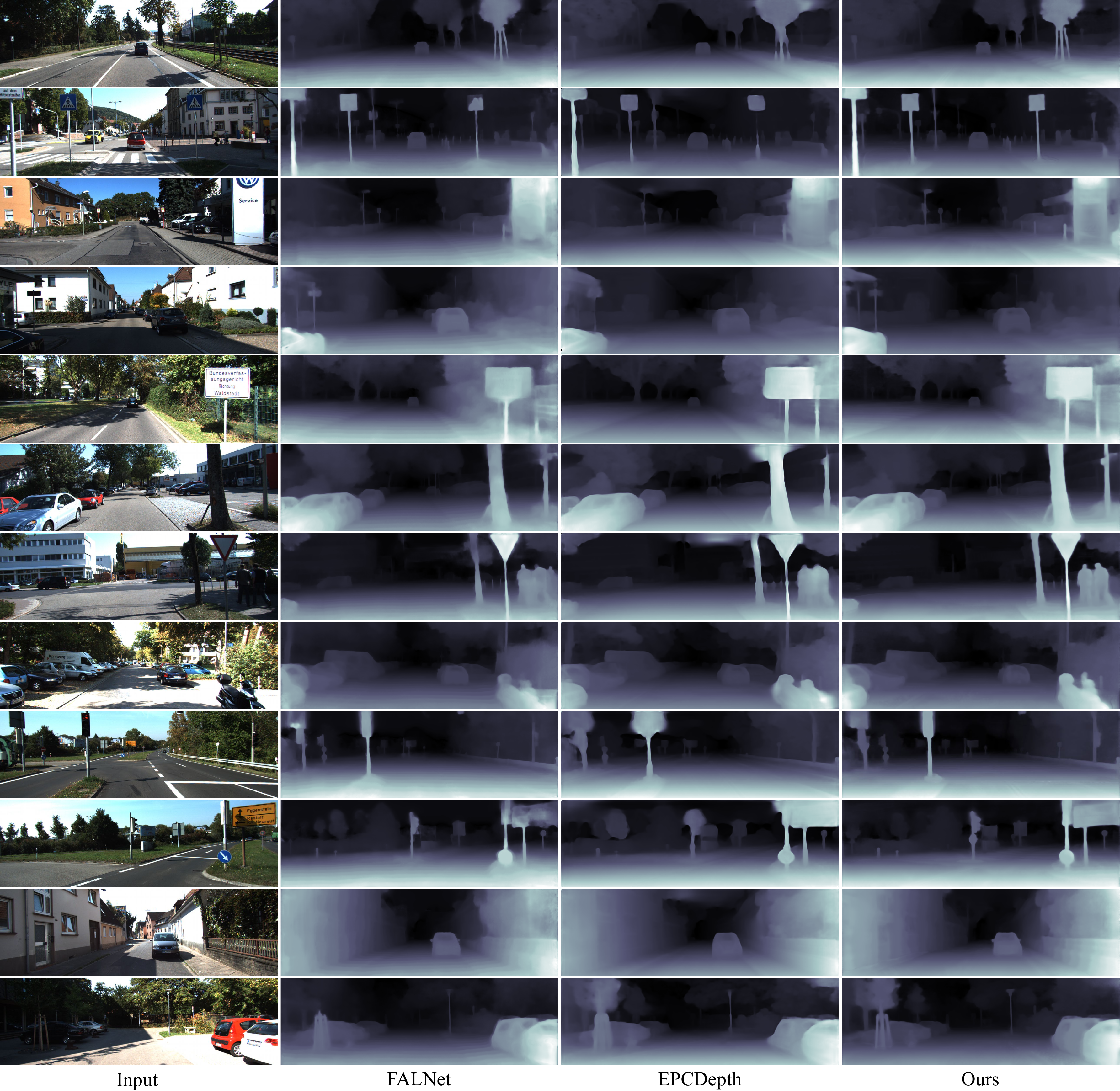}
\caption{\textbf{Additional qualitative results on the KITTI dataset.} Our network predicts smoother depth for the ground while preserving thin structures and sharp object edges with fewer depth artifacts.}
\label{sm_fig:final_compare_full}
\end{figure*}

\newpage

{\small
\bibliographystyle{ieee_fullname}
\bibliography{egbib}
}

\end{document}